\def\vepsilon{{\bm{\epsilon}}}

\documentclass{fairmeta}
\usepackage[most]{tcolorbox}

\usepackage{amsmath,amsfonts,bm}

\def\eqref#1{equation~\ref{#1}}

\def\1{\bm{1}}

\def\vzero{{\bm{0}}}

\def\va{{\bm{a}}}

\def\vc{{\bm{c}}}

\def\vf{{\bm{f}}}
\def\vg{{\bm{g}}}
\def\vh{{\bm{h}}}

\def\vl{{\bm{l}}}
\def\vm{{\bm{m}}}

\def\vo{{\bm{o}}}
\def\vp{{\bm{p}}}
\def\vq{{\bm{q}}}

\def\vs{{\bm{s}}}

\def\vy{{\bm{y}}}
\def\vz{{\bm{z}}}

\def\mA{{\bm{A}}}

\def\mI{{\bm{I}}}

\def\mQ{{\bm{Q}}}

\DeclareMathAlphabet{\mathsfit}{\encodingdefault}{\sfdefault}{m}{sl}
\SetMathAlphabet{\mathsfit}{bold}{\encodingdefault}{\sfdefault}{bx}{n}

\def\gA{{\mathcal{A}}}
\def\gB{{\mathcal{B}}}

\def\gD{{\mathcal{D}}}

\def\gI{{\mathcal{I}}}

\def\gL{{\mathcal{L}}}

\def\gN{{\mathcal{N}}}
\def\gO{{\mathcal{O}}}
\def\gP{{\mathcal{P}}}

\def\gS{{\mathcal{S}}}

\def\gV{{\mathcal{V}}}

\newcommand{\E}{\mathbb{E}}

\newcommand{\R}{\mathbb{R}}

\usepackage{amsmath}
\usepackage{amssymb}
\usepackage{bbm} 
\usepackage{url}
\usepackage{graphicx}
\usepackage{booktabs}
\usepackage{amsmath}
\usepackage{amsfonts}
\usepackage{wrapfig}
\usepackage{caption}
\usepackage{algorithm}
\usepackage{tabularx}
\usepackage{verbatim}
\usepackage{algpseudocode}
\usepackage{makecell}
\usepackage{array}
\usepackage{float}

\title{CoLA: Multi-VLA Coordination via Learned Adapters}
\newcommand{\codeurl}{\url{https://github.com/Ahaskar04/CoLA.git}}

\author[1,3,*]{Ahaskar Kashyap}
\author[2,3,*]{Jiawen Zong}
\author[1,3,*]{Boyu Ma}
\author[1,3]{Chuhao Zhou}
\author[1,3]{Jingliang Li}
\author[2,3]{\mbox{Stefano V. Albrecht}}
\author[1,3,\dagger]{Jianfei Yang}

\affiliation[1]{MARS Lab}
\affiliation[2]{Autonomous Agents Research Group}
\affiliation[3]{Nanyang Technological University, Singapore}

\contribution[*]{Equal contribution}
\contribution[\dagger]{Corresponding author}

\abstract{
Vision-Language-Action (VLA) models are typically deployed as isolated single-agent
policies. When several such agents share a workspace, they can coordinate implicitly,
by observing one another, or explicitly, by exchanging messages, but it remains
unclear when explicit communication becomes necessary. We introduce \textbf{CoLA} (\textbf{Co}ordination via \textbf{L}earned \textbf{A}dapters), which keeps each agent's VLA backbone frozen and trains a lightweight message channel and a per-arm diffusion action head on top, with fully decentralised execution. Across three tasks that differ in how much each agent can observe, the channel matters only when information is private. On an occluded tray handover, where the box's colour marker is visible to one agent only, CoLA reaches $76.0\%$ success versus $1.3\%$ with messages zeroed throughout training. On two-arm and three-arm handover tasks, removing messages has no measurable effect with a shared overhead camera ($90.6\%$ vs.\ $88.3\%$ at two arms; $63.5\%$ vs.\ $60.0\%$ at three), but lowers success on the two-arm task's wrist-only variant ($74.8\%$ vs.\ $64.3\%$). This architecture generalises across different VLA backbones as well. With a frozen $\pi_{0.5}$ backbone on the same occluded tray task, CoLA matches the overall success rate of a LoRA-fine-tuned baseline (37.3\% vs.\ 34.7\%) using a fraction of the trainable parameters. When CoLA places the box, it selects the correct tray $71.8\%$ of the time, whereas the independently fine-tuned baseline, whose placing arm never sees the marker, selects it at chance level ($34.7\%$ against $33.3\%$).
\footnote{Code: \codeurl}
}

\correspondence{\email{jianfei.yang@ntu.edu.sg}}

\begin{document}

\maketitle
\section{Introduction}
\label{sec:intro}

Robot foundation models trained on large cross-embodiment datasets now generalise well in single-arm manipulation~\citep{Brohan2022RT1, Zitkovich2023RT2, Ghosh2024Octo, Kim2024OpenVLA}. However, many real-world robotic deployments inherently require several arms or robots to act together, ranging from dual-arm collaborative handovers to multi-robot assembly and warehouse staging~\citep{krnjaic2024scalable, yifan2025embodied}. In such domains, task execution cannot rely exclusively on individual dexterity; success relies on mutual perception, intention signalling, and closed-loop behavioural synchrony between autonomous partners.

Despite their dexterity in isolation, current VLA models do not coordinate naturally when co-located in a shared workspace~\citep{dey2026autonomy}. Each operates under an egocentric assumption: it perceives the workspace through its own sensors and treats partner arms as unmodelled obstacles. Closing this gap faces three practical constraints. Training multi-arm policies end-to-end~\citep{rdt1b, Zhao2023ACT, Fu2024MobileALOHA} requires scarce multi-robot data and discards single-agent pretraining. LLM coordinators~\citep{Mandi2024RoCo, Liu2024COHERENT} run at $0.5$--$1\,\text{Hz}$, which is too slow for reactive manipulation. And deployed robots often run fixed on-device backbones whose weights are costly or impossible to modify (we defer further discussion of related works to \cref{sec:relatedwork}).

These constraints jointly define the setting we study, including agents whose backbones stay frozen, coordinating without a centralised arbiter, at real-time control rates. In this setting, agents can coordinate only implicitly, by observing one another, or through an explicit channel added alongside their frozen policies. We therefore ask: \textit{when does explicit inter-agent communication improve coordination over implicit observation alone, and how does this depend on task coupling and partial observability?}

To answer this question, we propose \textbf{Co}ordination via \textbf{L}earned \textbf{A}dapters (\textbf{CoLA}). CoLA discards each backbone's native action head entirely and in its place trains three modular components on top of each frozen VLA backbone: a latent message encoder, a partner message decoder, and a coordination-conditioned action chunking head. The encoder compresses the agent's readout embedding into a low-dimensional message vector. Each agent sends its encoded message to the other agents via the communication channel. Received messages are then decoded and fused with the agent's self representation, before being mapped to a coordinated action chunk. The encoders, decoders, and action heads are optimised jointly by behavioural cloning on multi-robot demonstrations.

We evaluate CoLA in four settings of increasing shared observability, built on three tasks: an occluded tray handover, where the colour marker is visible only to arm A and arm B must place the box in the matching tray; a two-arm handover with each
arm's own wrist camera only; the same two-arm handover with a shared overhead
camera; and a three-arm handover chain, which also shares the overhead camera. All
settings use a frozen Octo-Base backbone, and the tray task is additionally tested
with a frozen $\pi_{0.5}$ backbone. Against the strongest decentralised baseline in
each setting, head-only or fully fine-tuned, CoLA achieves between $5.5$ and $53.3$  percentage points higher success rate, with the largest gain on the occluded tray handover ($76.0\%$ vs.\ $22.7\%$). To isolate the channel itself, we retrain CoLA
with every message fixed to the zero vector during and after training. Removing the channel collapses success on the occluded tray task ($76.0\%$ to $1.3\%$) and lowers it under wrist-only
observation ($74.8\%$ to $64.3\%$), but has no measurable effect once the arms share
an overhead view ($90.6\%$ vs.\ $88.3\%$ at two arms and $63.5\%$ vs.\ $60.0\%$ at
three, both within seed-to-seed variation). Whether a team of frozen VLAs needs an
explicit channel is therefore determined by the task's information structure: the
channel is essential when task-relevant state is private to one agent, and redundant
when the agents can already observe it. Because the channel is a small module
trained on cached features of a frozen backbone, it can be added to already-deployed VLA policies without retraining them, and used only where the information structure calls for it. This makes the choice of when and what to communicate a design question in its own right for multi-robot VLA systems.

\section{Related Work}
\label{sec:relatedwork}

\textbf{Multi-Agent Reinforcement Learning and Emergent Communication.}
Learned inter-agent communication is well-studied in cooperative MARL~\citep{Albrecht2024MARL}. Architectures like CommNet~\citep{Sukhbaatar2016CommNet}, DIAL~\citep{foerster2016dial}, and TarMAC~\citep{Singh2018TarMAC} optimise continuous differentiable communication channels jointly with agent policies. Emergent communication shows that symbolic or continuous communication protocols can spontaneously ground coordination cues without direct linguistic supervision~\citep{Lazaridou2017Emergence, Havrylov2017Emergence, Baker2019EmergentTool}. Alternatively, communication can also be strictly bounded to the action space. Recent approaches, such as Multiagent Control via Action Suggestions (MCAS), demonstrate that sharing suggested joint actions, instead of full observations, allows agents to effectively infer belief states under decentralised partial observability~\citep{Asmar2024EfficientMP}. However, these methods train all agent policies from scratch. In CoLA, by contrast, each agent's feature extractor is a pretrained VLA that stays frozen while the channel is learned.

\textbf{Multi-Robot and Bimanual Manipulation.}
Prior multi-robot manipulation architectures typically rely on unified centralised controllers. Centralised imitation models such as ACT~\citep{Zhao2023ACT}, Mobile ALOHA~\citep{Fu2024MobileALOHA}, InterACT~\citep{Lee2024InterACT}, and RDT-1B~\citep{rdt1b} model multiple limbs within a single concatenated action space. TwinVLA~\citep{Chen2025TwinVLA} coordinates twin single-arm models via dense cross-attention, while AnyBimanual~\citep{Lu2025Anybimanual} transfers unimanual policies to bimanual control; both, however, update backbone weights. Closer to our setting, CHORUS~\citep{doshi2026chorusdecentralizedmultiembodimentcollaboration} adapts a single VLA backbone so that each robot runs an independent copy conditioned only on its own observations and a robot-identifying prompt, and MIMIC-D~\citep{dong2025mimic} jointly trains decentralised diffusion policies that coordinate implicitly from local information. Both deliberately avoid communication between agents at inference time; neither asks when such a channel becomes necessary, which is the question we study. LLM-based planners such as RoCo~\citep{Mandi2024RoCo} and COHERENT~\citep{Liu2024COHERENT} employ multi-agent text dialogues for task assignment, but their multi-second query latencies cannot sustain reactive physical handovers.

\textbf{Parameter-Efficient Fine-Tuning (PEFT).}
PEFT methods such as adapters~\citep{Houlsby2019Adapters}, LoRA~\citep{Hu2021LoRA}, and prompt tuning~\citep{Lester2021PromptTuning} adapt a single foundation model to a new domain, and in robotics are used to adapt VLAs to new embodiments or sensors~\citep{Kim2024OpenVLA, Kim2025FineTuningVLA}. CoLA instead repurposes adapters for communication between agents.

\section{Problem Formulation \& Methodology}
\label{sec:method}

\subsection{Decentralised Multi-Agent Formalisation}
\label{sec:formulation}

We model multi-robot cooperative manipulation as a decentralised partially observable process with communication, following the Dec-POMDP-Com~\citep{goldman2003communication} but without a reward function, as is standard in imitation learning~\citep{abbeel2004apprenticeship}. The model is defined by the tuple $\langle \gI, \gS, \{\gA_i\}, \Sigma, \gP, \{\gO_i\}, \Omega, \rho_0, T \rangle$. Here $\gI = \{1, \dots, N\}$ is the set of $N$ robot agents; $\gS$ is the global state space; $\gA_i$ is the continuous action space of agent $i$; and $\Sigma = \R^{d_m}$ is the space of messages agents may exchange. The transition distribution $\gP(\vs' \mid \vs, \va_1, \dots, \va_N)$ depends on physical actions only, so messages influence the environment solely through the actions they induce. Each agent $i$ receives observations from its local observation space $\gO_i$, drawn jointly according to $\Omega(\vo_1, \dots, \vo_N \mid \vs)$. Unlike the standard formulation, which conditions observations on the successor state and joint action, we condition on the current state only, since each observation is a camera rendering of that state. Finally, $\rho_0$ is the initial state distribution and $T$ is the episode horizon.

Under decentralised execution, each agent receives only its local observation $\vo_i^t \in \gO_i$ (an RGB camera stream), without direct access to partner observations $\vo_{j \neq i}^t$ or the global state $\vs^t$ (a subset of experiments relaxes this with a shared overhead view; those results are marked explicitly). Agents communicate over a channel graph defined by neighbourhood sets $\gN(i) \subseteq \gI \setminus \{i\}$: at each timestep $t$, agent $i$ broadcasts a message $\vm_i^t \in \Sigma$ and receives $\{\vm_j^t\}_{j \in \gN(i)}$. We restrict attention to \emph{reactive} policies, which condition on the current observation and the partners' current messages rather than on full action--observation histories:
\begin{equation}
\pi_i\big(\mA_i^t \;\big|\; \vo_i^t, \{\vm_j^t\}_{j \in \gN(i)}\big),
\label{eq:decentralised_policy}
\end{equation}
where $\mA_i^t = \big(\va_i^t, \va_i^{t+1}, \dots, \va_i^{t+K-1}\big) \in \R^{K \times D_a}$ denotes an open-loop chunk of $K$ future actions ($D_a = 7$). This restriction suffices when task-relevant state is either currently observable to some agent or recoverable from current proprioception.

Rather than optimising a reward, we are given demonstrations from a scripted expert team policy $\pi^E: \gS \to \gA_1 \times \dots \times \gA_N$, which maps the global state to a joint action $(\va_1^t, \dots, \va_N^t)$ for every agent at each timestep, so each demonstration trajectory records all $N$ agents' actions synchronously. The expert acts on the global state $\vs^t$, so it is not itself a decentralised reactive policy; a learner can reproduce its behaviour only to the extent that the required information is available through the agent's own observation or the messages it receives. Our goal is to learn decentralised reactive policies that imitate $\pi^E$, which we do by behavioural cloning (\cref{sec:objective}).

\subsection{Frozen Foundation Backbones}
\label{sec:backbones}
Each agent $i$ has a pretrained VLA backbone $f_\theta$ whose parameters remain frozen throughout training. At each timestep, the backbone encodes each of the agent's $V_i$ camera views $v \in \{1, \dots, V_i\}$, together with the task instruction $\vl$, into a feature vector, $\vf_i^{t,v} = f_\theta(\vo_i^{t,v}, \vl) \in \R^d$. The agent's proprioceptive state $\vq_i^t$ is mapped to an embedding $\vp_i^t = \phi_{\text{prop}}^i(\vq_i^t)$ by a small network trained jointly with the other modules. Because proprioception is easier to fit than the visual and message channels, a policy can shortcut through it and under-use those harder signals~\citep{lu2026h3dp}; to discourage this, $\vq_i^t$ is zeroed with probability $0.1$ during training. The agent's representation concatenates these:
\begin{equation}
\vh_i^t = \big[\, \vf_i^{t,v_1} \,\|\, \dots \,\|\, \vf_i^{t,v_{V_i}} \,\|\, \vp_i^t \,\big] \in \R^{d_h}.
\end{equation}
Both the CoLA message encoder and the action head take $\vh_i^t$ as input, so a message can carry information from any of the agent's views and from its physical state. Because $f_\theta$ is frozen, its features can be computed once and cached, and the trainable modules can only select and transmit information that $f_\theta$ already encodes.

\subsection{CoLA Architectural Design}
\label{sec:arch}

\begin{figure}[t]
    \centering
    \includegraphics[width=\linewidth]{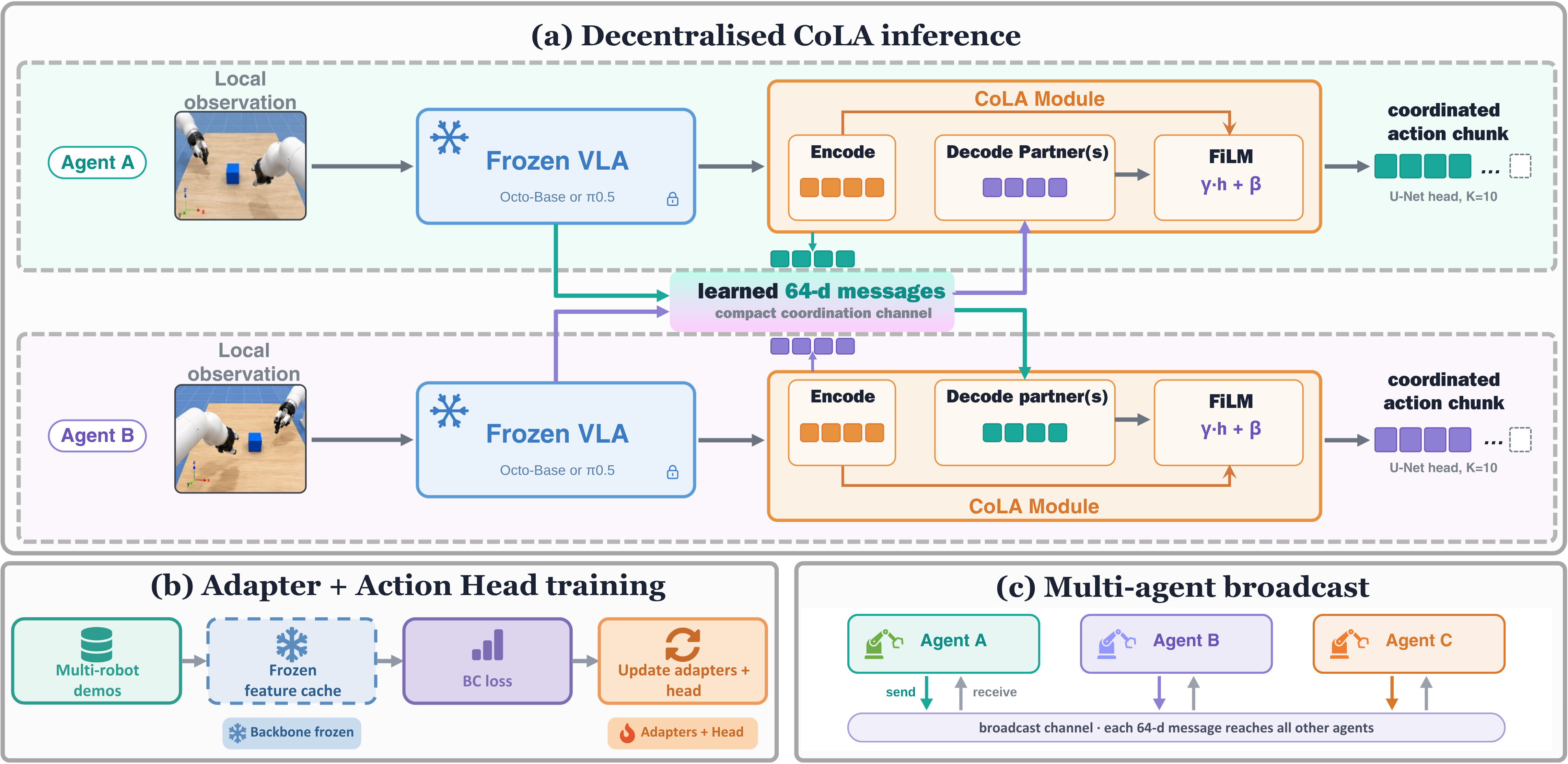}
    \caption{\textbf{CoLA framework overview.}
        (a)~\textit{Decentralised inference:} each agent's frozen VLA backbone (Octo-Base or $\pi_{0.5}$) encodes its local observation. Its CoLA module compresses the backbone feature into a compact message ($d_m = 64$), decodes the partner's message, and concatenates the result with its own representation to form $\vz_i$. This vector conditions a U-Net action head via FiLM, which predicts a coordinated action chunk ($K = 10$).
        (b)~\textit{Adapter and action-head training:} backbone features are cached offline and the backbones stay frozen. Only the CoLA modules are trained with a behavioural cloning loss, with gradients flowing end-to-end through the communication channel.
        (c)~\textit{Multi-agent broadcast:} each agent broadcasts its message to all other agents and receives theirs, so the same design scales to $N \geq 3$ agents.}
    \label{fig:cola_method}
\end{figure}

Each agent $i$ discards its backbone's native action head and instead hosts three trainable modules (\cref{fig:cola_method}):
\paragraph{Latent Message Encoder ($\phi_{\text{enc}}^i$).} A two-layer MLP with ReLU activations that compresses local backbone representations into a coordination message:
    \begin{equation}
    \vm_i^t = \phi_{\text{enc}}^i\big(\mathrm{sg}_{\vf}[\vh_i^t]\big) \in \R^{d_m}, \quad d_m = 64,
    \end{equation}
    where $\mathrm{sg}_{\vf}[\cdot]$ denotes a stop-gradient applied only to the backbone-feature components of $\vh_i^t$; the proprioceptive encoder $\phi_{\text{prop}}^i$ receives gradients through the message.

\paragraph{Message Decoder ($\phi_{\text{dec}}^{j \to i}$).} One decoder per
\emph{ordered pair}, mapping partner $j$'s message into $i$'s feature space,
\begin{equation}
\vc_{j \to i}^t = \phi_{\text{dec}}^{j \to i}(\vm_j^t) \in \R^d, \quad \forall j \in \gN(i).
\end{equation}
Decoders are not shared across senders: $i$ learns a distinct mapping for each partner, so that in the three-arm task B reads A's and C's messages through separate weights.

\paragraph{Coordination Action Head.} Concatenates the agent's own representation with its decoded partner embeddings,
\begin{equation}
\vz_i^t = \big[\,\vh_i^t \;\big\|\; \{\vc_{j \to i}^t\}_{j \in \gN(i)}\,\big]
\in \R^{d_h + |\gN(i)|\,d},
\end{equation}
which conditions the diffusion head of \cref{sec:objective}: a 1D temporal U-Net with FiLM conditioning that predicts the noise added to the joint
chunk, alongside a logit head for the binary gripper column.

\paragraph{Action Chunking Dynamics.}
Actions are generated in chunks of $K = 10$ steps executed open-loop at $50\,\text{Hz}$ control rate ($0.2\,\text{s}$ per chunk). The policy is queried once per chunk, giving a $5\,\text{Hz}$ inference rate (one query every $200\,\text{ms}$).

\subsection{Multi-Agent Topology Routing}
\label{sec:routing}

CoLA extends to $N$ agents through flexible graph topologies:
\paragraph{Bimanual Teams ($N=2$):} Direct peer-to-peer exchange, $\gN(1)=\{2\}$ and $\gN(2)=\{1\}$. Each agent's head input is $[\vh_i \,\|\, \vc_{j\to i}]$, of dimension $d_h + d$.

\paragraph{Tri-Manual System ($N=3$):} A fully connected topology, $\gN(i) = \gI \setminus \{i\}$ for every agent, giving six directed channels. Each agent decodes \emph{both} partners' messages through its own per-sender decoder, $\vz_i = [\vh_i \,\|\, \vc_{j\to i} \,\|\, \vc_{k\to i}]$ of dimension $d_h + 2d$. The architecture is symmetric: no agent acts as a hub, all three heads take identically shaped input, and encoder and decoder architectures are identical across agents.

\subsection{End-to-End Imitation Learning Objective}
\label{sec:objective}

Given a multi-agent demonstration dataset $\gD = \{(\vo_{1:N}^t, \mA_{1:N}^t)\}$ collected from scripted experts, CoLA is trained by behavioural cloning. We split each expert chunk as $\mA_i^t = [\mQ_i^t, \tilde{\vg}_i^t]$, where $\mQ_i^t \in \R^{K \times 6}$ holds the joint targets and $\tilde{\vg}_i^t \in \{-0.9, 0.9\}^K$ the normalised gripper channel, from which we derive binary labels $\vy_i^t = \mathbbm{1}[\tilde{\vg}_i^t > 0] \in \{0,1\}^K$ for the classification loss. Following diffusion policies~\citep{ho2020denoising,chi2025diffusion}, we sample, independently for each agent, a diffusion step $\tau_i$ uniformly from $\{1, \dots, S-1\}$ with $S = 100$, and noise $\vepsilon_i \sim \mathcal{N}(\vzero, \mI)$. We then form the corrupted joint chunk $\tilde{\mQ}_i^t = \sqrt{\bar\alpha_{\tau_i}}\,\mQ_i^t + \sqrt{1-\bar\alpha_{\tau_i}}\,\vepsilon_i$ under a cosine schedule~\citep{nichol2021improved}, and train each agent's head $\phi_{\text{unet}}^i$ to predict the added noise conditioned on $\vz_i^t$:
\begin{equation}
\gL(\boldsymbol\phi) = \E_{\gD}\Big[\sum_{i=1}^{N} \E_{\tau_i,\,\vepsilon_i}\big[\gL_i\big]\Big],
\qquad
\gL_i = \frac{1}{6K}\big\|\phi_{\text{unet}}^i(\tilde{\mQ}_i^t, \tau_i, \vz_i^t) - \vepsilon_i\big\|_F^2
+ \lambda\,\mathrm{BCE}\big(\phi_{\text{grip}}^i(\vz_i^t),\, \vy_i^t\big),
\label{eq:bc_loss}
\end{equation}
where $\phi_{\text{grip}}^i$ outputs $K$ gripper logits, the BCE is averaged over the $K$ steps, and $\lambda = 0.25$. At test time, we sample with DDIM~\citep{song2020denoising} using $10$ denoising steps. We predict the gripper with a separate classification head because it is binary: in early development, predicting it as a continuous dimension produced near-constant outputs, which motivated the separate head. The classification head reaches $98.7$--$100\%$ validation gripper accuracy across all four coordination regimes (\cref{tab:gripper_acc}).

\textbf{Gradients Through the Channel.}
Because $\vz_i^t$ depends on partners' messages, gradients flow across the channel. Agent $j$'s message $\vm_j^t$ enters only its receivers' action heads, so $j$'s encoder is trained entirely by its partners' losses, whereas the decoder producing $\vc_{j \to i}^t$ belongs to agent $i$ and is trained by $\gL_i$ alone. Each encoder is therefore pushed to transmit whatever its partners need to disambiguate their actions.

\textbf{Offline Feature Caching.}
Because $f_\theta$ is frozen, its features are computed once per dataset and reused by every run and ablation, so training CoLA
takes under one GPU-hour, against $2$--$49$ for the fine-tuned baselines at the same
optimiser budget (\cref{tab:training_cost}).


\section{Experiments}
\label{sec:experiments}

Our experiments aim to answer the following questions: 
\textbf{(Q1)} Can agents built on frozen VLA backbones coordinate under decentralised execution when only a message channel and a new action head are trained?
\textbf{(Q2)} How much does the message channel contribute to success? 
\textbf{(Q3)} Does the approach generalise to three agents and to different frozen backbones?

\subsection{Tasks}
\label{sec:tasks}
\begin{figure}[ht]
    \centering
    \includegraphics[width=1\linewidth]{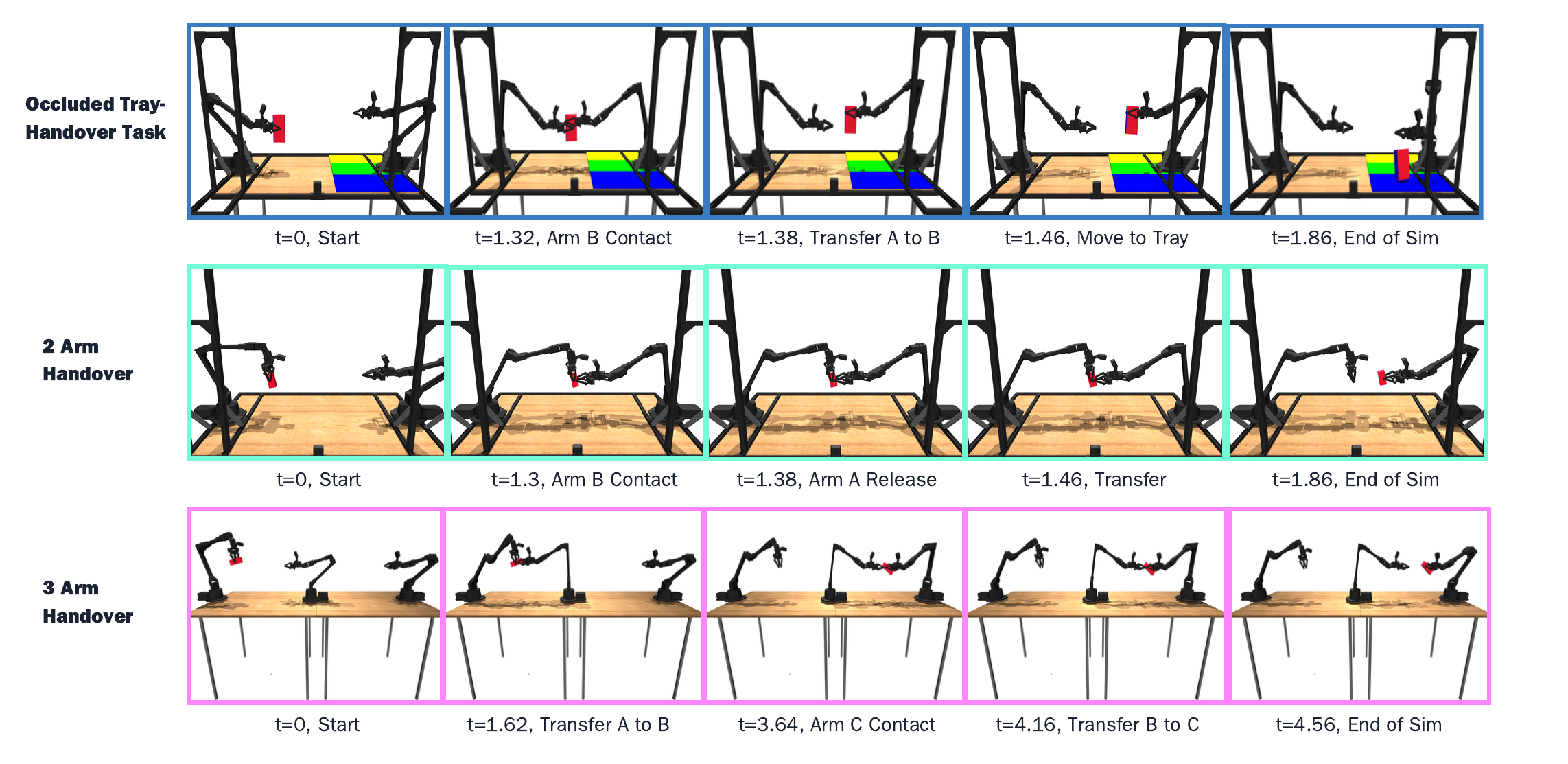}
    \caption{\textbf{Multi-Agent Manipulation Progression.} Key milestones across evaluated tasks: (top) Occluded Tray-Handover Task, (middle) 2-arm Reactive Handover, and (bottom) 3-arm Handover Chain.}
    \label{fig:tasks}
\end{figure}

We evaluate on three tasks that differ in how much task-relevant information each agent can observe with its own cameras (\cref{fig:tasks}).

\paragraph{Occluded tray handover.} Arm A grasps a box bearing one of three coloured markers (blue, green, or yellow) on the face pointing toward Arm A, and hands it to Arm B, which must place it on the tray of the matching colour on its own side of the table. Each arm observes only its own wrist camera. The marker appears in Arm A's camera in $100\%$ of control steps and in Arm B's in under $5\%$. Without communication, Arm B therefore has almost no access to the colour, and among the episodes in which it places the box it can expect to select the correct tray only about one time in three. An episode is successful if the box enters the correct tray.

\paragraph{Two-arm handover.} Arm A starts holding the object and presents it at a handoff point; Arm B must approach, grasp, and hold while A releases. A's presentation pose is randomised by $\pm 10\,\text{cm}$ laterally and vertically. The challenge is timing: B must close its gripper before A opens its own, and a premature release drops the object. We evaluate two variants: each arm's wrist camera only, and each arm's wrist camera together with a shared overhead camera. An episode is successful if Arm B grasps the object without dropping it.

\paragraph{Three-arm handover chain.} The object passes A $\to$ B $\to$ C, with A's presentation randomised as above, and all arms additionally share the overhead view alongside their own wrist camera. B must receive the object and pass it on, C's approach depends on when B is ready, and a failure at either transfer ends the episode. An episode is successful if both transfers complete and Arm C then holds the object for $20$ steps.

\subsection{Experimental Setup}
\label{sec:setup}

\paragraph{Simulation Environments.}
All environments are simulated in MuJoCo~\citep{todorov2012mujoco} with a $2\,\text{ms}$ physics timestep ($500\,\text{Hz}$) and a control decimation of $10$, giving a $50\,\text{Hz}$ control rate. We control $N \in \{2,3\}$ 6-DoF Trossen ViperX~300 arms with parallel-jaw grippers. Each arm's action is $7$-dimensional: target positions for its six revolute joints, normalised per dimension to $[-0.9, 0.9]$ using min--max statistics from the training split, plus a gripper command, which CoLA predicts as a binary open/closed decision. Each agent observes a $256 \times 256$ RGB image from its own wrist-mounted camera, together with its $7$-dimensional proprioceptive state (six joint angles and gripper opening). In the overhead variants of the two- and three-arm handover, every agent additionally observes a shared fixed overhead camera alongside its own wrist camera (\cref{sec:formulation}), for CoLA and the baselines alike. The occluded tray-handover task uses wrist cameras only.

All three tasks use Octo-Base ($93$M parameters). On the occluded tray-handover task we additionally evaluate $\pi_{0.5}$ (base checkpoint, ${\sim}3.3$B parameters~\citep{intelligence2025pi_}). Both backbones are frozen in their entirety and run once offline, encoding each agent's observation at each timestep into a single cached feature vector: the $768$-dimensional readout embedding for Octo-Base and, because $\pi_{0.5}$ has no readout token, a padding-masked mean over its final-layer PaliGemma prefix tokens (image, prompt, and joint state as $256$-bin tokens, matching the LoRA baseline) giving a $2048$-dimensional vector. On this backbone, the $7$-d proprioceptive state therefore enters CoLA twice: once inside the frozen $2048$-D feature (via the prompt tokens) and again through $\phi_{\text{prop}}$ (\cref{sec:backbones}); the proprio dropout described there regularises only the latter copy. Neither backbone's own action head is used. CoLA's modules (\cref{sec:arch}) have the same architecture across backbones; only the layer widths that depend on the backbone feature dimension change ($14.5$M trainable parameters with Octo-Base, $16.9$M with $\pi_{0.5}$, of which the communication channel accounts for $0.33$M and $0.83$M respectively and the remainder is the per-arm action head and proprioceptive encoder). CoLA uses a single generic instruction, ``coordinate with partner'', for every arm and task, whereas the baselines receive role-specific instructions (\cref{tab:instructions}). No instruction names the marker colour.
Pretrained Octo does not complete the two-arm handover zero-shot with its own action head (\cref{tab:zero_shot}). 

\paragraph{Evaluation Protocol.}
For the handover tasks, each policy is evaluated on $200$-episode blocks at evaluation-seed offsets drawn from $0$, $1000$, $2000$, $3000$ and $4000$, and we report the mean $\pm$ standard deviation of success rate across the blocks evaluated. The number of blocks differs by cell: five for every method on the 2-arm overhead task except CoLA with zeroed messages (three); three for every method on the 2-arm wrist-only task (offsets $0$, $1000$, $2000$); and five for the baselines but a single block (offset $0$) for both CoLA rows on the three-arm chain. The per-seed blocks behind every cell are given in \cref{tab:ho_overhead_seeds,tab:ho_wrist_seeds,tab:3arm_seeds}. The occluded tray-handover task keeps the handover positions fixed and randomises only hidden marker colour among three discrete values. A single seed with $150$ episodes already samples each colour densely (${\sim}50$ episodes per colour), therefore this task is evaluated with only a single seed ($0$). For this task we report success rate with a $95\%$ Wilson confidence interval over episodes, with the per-stage breakdown in \cref{tab:tray_blocks}.

\paragraph{Baselines for comparison.} All baselines are trained on the same demonstrations and episode splits as CoLA and, unless stated otherwise, on the same Octo-Base backbone. Within each task, every method uses the same number of optimiser steps at the same effective batch size of $128$ ($12{,}600$, $20{,}250$ and $26{,}850$ steps for the two-arm handover, tray and three-arm handover tasks, respectively). This matches the optimisation budget, not compute or trainable capacity. Learning rates and optimisers follow each method's standard recipe (Adam without weight decay for CoLA; AdamW with weight decay $10^{-4}$ for the Octo baselines), and full architectural and optimisation settings are listed in \cref{tab:system_params} (\cref{app:budget}). Since CoLA trains on cached backbone features, matching optimiser steps gives the fine-tuned baselines substantially more compute than CoLA. 

\begin{itemize}
    \item \textbf{Independent decentralised policies:} one policy per arm, conditioned on its wrist camera on the wrist-only tasks, or on both its wrist camera and the shared overhead camera on the overhead tasks, trained either (a)~\emph{head-only}, with the backbone frozen and a freshly initialised action head, or (b)~with \emph{full fine-tuning} of all backbone weights on the multi-robot demonstrations. All Octo baselines receive proprioception through Octo's proprio tokenizer: each decentralised policy sees its own arm's 7-D joint state, and the centralised policy both arms' (14-D). Because Octo-Base was pretrained without proprioception, this projection is newly initialised; head-only baselines keep it frozen at that initialisation, while full fine-tuning trains it.
    
    \item \textbf{Centralised policy:} a single model that outputs the joint 14-D action, in the same two regimes, from both arms' wrist cameras on the wrist-only tasks, or from both arms' wrist cameras plus the shared overhead camera on the overhead tasks. It violates decentralised execution and serves as a privileged reference.
    
    \item \textbf{CoLA with zeroed messages:} the full CoLA architecture, trained and evaluated with all messages set to zero. It has the same architecture, parameter count, and training recipe as CoLA; the only difference is that the channel carries no information, which isolates the contribution of communication.
    
    \item \textbf{Cross-backbone transfer (occluded tray-handover task):} independent per-arm $\pi_{0.5}$ policies fine-tuned with openpi's LoRA recipe (LoRA on both Gemma experts plus the full vision encoder; $467.0$M trainable parameters per arm), compared against CoLA on the frozen $\pi_{0.5}$ backbone ($8.5$M per arm), testing whether CoLA depends on a particular backbone.
    
\end{itemize}

\subsection{Main Empirical Results}
\label{sec:main_results}

\begin{table}[htb!]
\centering
\setlength{\tabcolsep}{4pt}
\caption{\textbf{Main results.} Success rate (\%), ordered from least to most shared observability. HO = handover. Handover columns: mean $\pm$ std over 200-episode evaluation blocks (\cref{sec:setup}); every row is a single training run. Tray: single 150-episode block. --: not evaluated.}
\label{tab:main_results}
\begin{tabular}{lcccc}
\toprule
\textbf{Method} & \makecell[c]{\textbf{Occluded}\\\textbf{Tray-HO}\\\textbf{(wrist only)}} & \makecell[c]{\textbf{2-Arm HO}\\\textbf{(wrist only)}} & \makecell[c]{\textbf{2-Arm HO}\\\textbf{(overhead)}} & \makecell[c]{\textbf{3-Arm HO}\\\textbf{(overhead)}} \\
\midrule
CoLA (with message)           & \textbf{76.0}              & 74.8 $\pm$ 2.6             & 90.6 $\pm$ 1.7             & \textbf{63.5}$^\dagger$    \\
CoLA (message zeroed)         & 1.3                        & 64.3 $\pm$ 6.7             & 88.3 $\pm$ 2.9             & 60.0$^\dagger$             \\
Decentralised, head-only      & 22.7                       & 43.8 $\pm$ 2.0             & 68.3 $\pm$ 2.6             & 20.5 $\pm$ 2.7             \\
Centralised, head-only        & 16.0                       & 21.0 $\pm$ 2.2             & 65.9 $\pm$ 4.7             & ---                        \\
\midrule
Centralised, full fine-tune   & 10.7                       & \textbf{80.0 $\pm$ 2.6}    & \textbf{92.1 $\pm$ 2.0}    & ---                        \\
Decentralised, full fine-tune & 12.0                       & 69.3 $\pm$ 1.8             & 82.4 $\pm$ 3.1             & 31.6 $\pm$ 4.5             \\
\bottomrule
\end{tabular}

\vspace{1mm}
\small
$^\dagger$ single 200-episode block. Tray chance is 33.3\% conditional on reaching a tray. Per-seed and per-stage results: \cref{app:results}.
\end{table}

\textbf{Q1: Coordination Efficacy.}
\cref{tab:main_results} shows that CoLA outperforms every decentralised baseline on all four tasks, while modifying only lightweight adapters and an action head atop a frozen backbone. On the occluded tray-handover task, independent decentralised policies perform poorly, while CoLA reaches $76.0\%$. On the handover tasks, CoLA exceeds the decentralised head-only baseline by $31.0$ percentage points on wrist-only handover ($74.8\%$ vs.\ $43.8\%$), $22.3$ percentage points with a shared overhead view ($90.6\%$ vs.\ $68.3\%$), and $43.0$ points at three arms ($63.5\%$ vs.\ $20.5\%$). CoLA also approaches the privileged centralised full fine-tune on handover ($90.6\%$ vs.\ $92.1\%$ with a shared view; $74.8\%$ vs.\ $80.0\%$ wrist-only) while retaining fully decentralised execution. \cref{fig:qual_handover} contrasts a CoLA rollout with an uncoordinated baseline, and \cref{sec:failure_modes} categorises the failures.

\textbf{Q2: Is Communication Causally Necessary?}
The head-only baseline underperforms no-message CoLA on every handover task (e.g., $68.3\%$ vs.\ $88.3\%$ with a shared overhead view). Additionally, CoLA replaces Octo's action head with a diffusion U-Net with FiLM conditioning, predicts the gripper with a separate classification head, adds a learned proprioceptive embedding, and uses a different chunk length and training schedule. Comparisons against the Octo baselines therefore measure CoLA's whole pipeline, not communication. The no-message ablation isolates the channel: it keeps CoLA's architecture and training recipe unchanged and removes only the message content. Removing messages collapses the occluded tray handover ($76.0\%$ to $1.3\%$), although most of this drop occurs before tray selection: the message-zeroed model completes the handover itself in only $8.0\%$ of episodes, against $88.7\%$ for the head-only baseline, which also has no channel (\cref{tab:tray_blocks}). Training with a constant zero message therefore impairs the whole policy on this task, so $1.3\%$ is not just the cost of withholding the colour alone. The evidence that the channel carries the colour is tray choice given a placement ($85.7\%$ for CoLA against $26.4\%$ for the head-only baseline, chance $33.3\%$) and the probes of \cref{sec:probing}. Removing messages also degrades the wrist-only handover ($74.8\%$ to $64.3\%$), but has no measurable effect when a shared overhead view shows each agent its partner ($90.6\%$ vs.\ $88.3\%$ at two arms; $63.5\%$ vs.\ $60.0\%$ at three, within noise). The channel therefore matters to the extent that task-relevant information is private to one agent. When agents share a view, CoLA's advantage over the independent baselines comes from the pipeline differences above, which we do not isolate individually. \cref{tab:tray_ablations} shows that performance is largely insensitive to message width $d_m \in \{16, 64, 128\}$.

\begin{table}[t]
\centering
\small
\setlength{\tabcolsep}{4pt}
\caption{\textbf{Cross-backbone transfer on the occluded tray-handover task.} CoLA on a frozen $\pi_{0.5}$ backbone versus independently LoRA-fine-tuned per-arm $\pi_{0.5}$ policies. \emph{Placed}: the box reaches any tray. \emph{Correct $\mid$ placed}: the box reaches the correct tray, given that it reaches one (chance $33.3\%$). \emph{Overall}: the box reaches the correct tray, over all episodes. Brackets give 95\% Wilson intervals; \textbf{bold} marks a value whose interval does not overlap the other method's.}
\label{tab:pi05_transfer}
\begin{tabular}{lccccc}
\toprule
\textbf{Method} & \makecell[c]{\textbf{Params}\\\textbf{(per arm)}}
                & \makecell[c]{\textbf{Train}\\\textbf{(GPU-h)}}
                & \makecell[c]{\textbf{Placed}\\\textbf{(\%)}}
                & \makecell[c]{\textbf{Correct $\mid$ placed}\\\textbf{(\%)}}
                & \makecell[c]{\textbf{Overall}\\\textbf{(\%)}} \\
\midrule
CoLA & $8.5$M & $0.2$
 & \makecell[c]{52.0\\{\scriptsize[44.1, 59.8]}}
 & \makecell[c]{\textbf{71.8}\\{\scriptsize[61.0, 80.6]}}
 & \makecell[c]{37.3\\{\scriptsize[30.0, 45.3]}} \\
\addlinespace[2pt]
Decentralised, LoRA fine-tune & $467.0$M & $24.4$
 & \makecell[c]{\textbf{100.0}\\{\scriptsize[97.5, 100]}}
 & \makecell[c]{34.7\\{\scriptsize[27.5, 42.6]}}
 & \makecell[c]{34.7\\{\scriptsize[27.5, 42.6]}} \\
\bottomrule
\end{tabular}
\end{table}

\textbf{Q3: Does CoLA's Advantage Persist with Team Size and Backbone?}
At three arms, CoLA outperforms decentralised full fine-tuning by $31.9$ percentage points ($63.5\%$ vs.\ $31.6\%$), up from $8.2$ percentage points at two arms ($90.6\%$ vs.\ $82.4\%$), and the head-only baseline by $43.0$ points (up from $22.3$). Because zeroed-message CoLA reaches $60\%$, this widening margin reflects CoLA's architecture rather than communication when an overhead view is shared; whether the channel's own contribution grows with team size requires a three-arm task with private information. Per-stage transfer rates for the chain are reported in \cref{tab:3arm_stages}.
 
To test whether the channel's benefit is specific to Octo, we repeat the occluded tray-handover task with a frozen $\pi_{0.5}$ backbone (\cref{tab:pi05_transfer}). The two methods fail in different places. The LoRA-fine-tuned per-arm policies place the box on a tray in every episode but choose the correct tray at chance level ($34.7\%$ against a chance rate of $33.3\%$), consistent with the placing arm having no access to the hidden colour. CoLA places the box in only $52.0\%$ of episodes, but when it does, it selects the correct tray $71.8\%$ of the time (95\% CI $[61.0, 80.6]$), well above both chance and the baseline. The message channel therefore still carries the hidden task information on $\pi_{0.5}$, what does not transfer is manipulation reliability, and overall success is statistically indistinguishable ($37.3\%$ vs.\ $34.7\%$).

\subsection{Probing What the Channel Encodes}
\label{sec:probing}

We analyse the information transmitted in the message by fitting linear probes (multinomial logistic regression for colour, ridge regression for pose) from three representations of the sending agent: its own features $\vh_i^t$ (\emph{self}), its message $\vm_i^t$ from the trained checkpoint (\emph{message}), and the encoder output from the message-zeroed checkpoint (\emph{control}), which receives no gradient when messages are zeroed (\cref{sec:objective}) and so stays at random initialisation. On the occluded tray-handover task, the message decodes occluded box colour at $73.3\%$ versus chance-level $33.3\%$ for the control (episode-level majority vote; per frame, $44.2\%$ vs.\ $39.7\%$), a $40$-point gap attributable to training through the channel that corroborates the necessity results (\cref{fig:probes} and \cref{tab:probes}, \cref{app:results}).

\section{Conclusion and Limitations}
\label{sec:conclusion}

\textbf{Conclusion.} We presented CoLA, a framework that trains a message-passing channel and a new action-generation pipeline on top of frozen, independently deployed VLA backbones, and used it to measure when that channel is actually necessary for coordination. Isolating the channel's effect from the rest of the architecture, we found communication is not uniformly beneficial. It is necessary when task-relevant state is private to one agent, and has no measurable effect once agents share a view of the workspace. On the occluded tray-handover task, the channel also carried the hidden colour with a frozen $\pi_{0.5}$ backbone, although overall success there did not exceed that of the LoRA-fine-tuned baseline. Our results suggest that explicit communication and implicit, shared-context coordination are not competing solutions but complementary ones, each suited to a different information regime, and that determining which regime a task falls into is itself a necessary step in designing multi-agent VLA systems.

\textbf{Limitations.} All experiments run in simulation on scripted expert
demonstrations, and we have not tested transfer to physical robots. CoLA also
assumes zero-latency synchronous communication. Although realistic for local
high-speed bus architectures, performance under asynchronous network dropouts and
variable latency remains to be benchmarked. Additionally, CoLA connects agents with
a fully connected topology and a separate decoder for every ordered pair
(\cref{sec:routing}), so both the number of channels and the number of decoders grow
as $O(N^2)$, with six of each at $N=3$. This is practical at the team sizes we
study, but scaling to dozens of robots will require sharing decoder weights across
senders, or aggregating messages through graph neural network or cross-attention
routers~\citep{Singh2018TarMAC}.

\newpage

\subsection*{Reproducibility Statement}
Environments use MuJoCo 3.4.0 with the Trossen ViperX 300s models from MuJoCo Menagerie (physics 500\,Hz, control 50\,Hz). All demonstrations are successful expert episodes, split 80/10/10 by episode. Frozen Octo-Base features (and Octo-Small, used only in \cref{tab:zero_shot}) are pre-extracted in JAX and cached. Hyperparameters are listed in \cref{tab:system_params} and the full procedure in \cref{alg:training}. Experiments ran on one node with 8 NVIDIA A100 (40\,GB) GPUs. Code: \codeurl.

\subsection*{AI Use Statement}
Generative AI tools were used exclusively for rephrasing and condensing author written text, formatting assistance, spelling and grammar refinement. All conceptual formulations, experimental analyses, and technical claims were designed and verified by the authors.

\clearpage
\newpage
\bibliographystyle{plainnat}
\bibliography{paper}

\clearpage
\appendix

\phantomsection
\section*{Appendix}

\begin{itemize}
  \item[] \textbf{\ref{app:training} \nameref{app:training}}

  \item[] \textbf{\ref{app:budget} \nameref{app:budget}}
    \begin{itemize}
      \item[] \ref{app:budget_hyperparams} \nameref{app:budget_hyperparams}
      \item[] \ref{app:budget_instructions} \nameref{app:budget_instructions}
      \item[] \ref{app:budget_cost} \nameref{app:budget_cost}
    \end{itemize}

  \item[] \textbf{\ref{app:figures} \nameref{app:figures}}
    \begin{itemize}
      \item[] \ref{app:figures_qual} \nameref{app:figures_qual}
    \end{itemize}

  \item[] \textbf{\ref{app:results} \nameref{app:results}}
    \begin{itemize}
      \item[] \ref{app:results_2arm} \nameref{app:results_2arm}
      \item[] \ref{app:results_3arm} \nameref{app:results_3arm}
      \item[] \ref{app:results_tray} \nameref{app:results_tray}
      \item[] \ref{app:results_probes} \nameref{app:results_probes}
      \item[] \ref{app:results_zeroshot} \nameref{app:results_zeroshot}
      \item[] \ref{app:results_gripper} \nameref{app:results_gripper}
    \end{itemize}

  \item[] \textbf{\ref{app:training_curves} \nameref{app:training_curves}}

  \item[] \textbf{\ref{sec:failure_modes} \nameref{sec:failure_modes}}
    \begin{itemize}
      \item[] \ref{sec:failure_upstream} \nameref{sec:failure_upstream}
      \item[] \ref{sec:failure_temporal} \nameref{sec:failure_temporal}
      \item[] \ref{sec:failure_posttransfer} \nameref{sec:failure_posttransfer}
    \end{itemize}
\end{itemize}

\clearpage

\section{Training Algorithm}
\label{app:training}

\begin{algorithm}[H]
\caption{CoLA Optimisation with Offline Feature Caching (diffusion U-Net head)}
\label{alg:training}
\begin{algorithmic}
\Require Multi-agent demonstration dataset $\gD = \{(\vo_{1:N}^t, \vq_{1:N}^t, \va_{1:N}^t)\}_{t=1}^T$ (images, proprioception, actions); frozen VLA backbone $f_\theta$ with instruction $\vl$; chunk horizon $K=10$; message dimension $d_m=64$; diffusion steps $S=100$ with cosine schedule $\{\bar\alpha_\tau\}_{\tau=0}^{S-1}$; gripper weight $\lambda=0.25$; proprio dropout $p=0.1$.
\State \textbf{// Phase 1: One-Time Offline Feature Extraction}
\State Split $\gD$ by episode into $\gD_{\text{train}}, \gD_{\text{val}}, \gD_{\text{test}}$ (80/10/10).
\For{each split $\gD_s \in \{\gD_{\text{train}}, \gD_{\text{val}}, \gD_{\text{test}}\}$}
    \For{each batch of multi-agent observations $\vo_{1:N}$}
        \State $\vf_i \leftarrow \mathrm{sg}\big[f_\theta(\vo_i, \vl).\texttt{readout\_action}[:,-1,0,:]\big], \quad \forall i \in \{1,\dots,N\}$
        \State Cache $(\vf_{1:N}, \vq_{1:N}, \va_{1:N})$ to disk.
    \EndFor
\EndFor
\State \textbf{// Phase 2: Action Chunk Indexing}
\State Index valid starting timesteps $\gV = \{t \mid [t, t+K-1] \text{ belongs to the same episode}\}$.
\State \textbf{// Phase 3: Adapter Parameter Optimisation}
\State Initialise $\boldsymbol{\phi} = \{\phi_{\text{prop}}^i, \phi_{\text{enc}}^i, \{\phi_{\text{dec}}^{j \to i}\}_{j \in \gN(i)}, \phi_{\text{unet}}^i, \phi_{\text{grip}}^i\}_{i=1}^N$; $\text{best\_val\_loss} \leftarrow \infty$.
\For{epoch $= 1$ to $150$}
    \For{minibatch $\gB \subset \gV_{\text{train}}$ of size 128}
        \State Load cached $\{\vf_i^t, \vq_i^t\}_{i=1}^N$ and target chunks $\{\mA_i^t = \va_i^{t:t+K-1}\}_{i=1}^N$.
        \For{agent $i = 1$ to $N$}
            \State Drop proprioception: $\tilde\vq_i^t \leftarrow b_i\,\vq_i^t,\; b_i \sim \mathrm{Bernoulli}(1-p)$.
            \State Self representation: $\vh_i^t \leftarrow [\mathrm{sg}[\vf_i^t] \,\|\, \phi_{\text{prop}}^i(\tilde\vq_i^t)]$.
            \State Encode message: $\vm_i^t \leftarrow \phi_{\text{enc}}^i(\vh_i^t)$.
        \EndFor
        \For{agent $i = 1$ to $N$}
            \State Decode partner messages: $\vc_{j \to i}^t \leftarrow \phi_{\text{dec}}^{j \to i}(\vm_j^t), \; \forall j \in \gN(i)$.
            \State Conditioning vector: $\vz_i^t \leftarrow [\vh_i^t \,\|\, \{\vc_{j \to i}^t\}_{j \in \gN(i)}]$.
            \State Split target: joint chunk $\mA_i^{\text{jnt}} \in \R^{K \times 6}$, gripper labels $\vy_i \leftarrow \mathbb{1}[\tilde\vg_i^t > 0] \in \{0,1\}^K$.
            \State Sample $\tau_i \sim \mathcal{U}\{1,\dots,S-1\}$ and $\vepsilon_i \sim \mathcal{N}(\mathbf{0}, \mI)$, independently per agent.
            \State Noise the joints: $\tilde\mA_i \leftarrow \sqrt{\bar\alpha_{\tau_i}}\,\mA_i^{\text{jnt}} + \sqrt{1-\bar\alpha_{\tau_i}}\,\vepsilon_i$.
            \State Predict noise: $\hat\vepsilon_i \leftarrow \phi_{\text{unet}}^i(\tilde\mA_i, \tau_i, \vz_i^t)$ \Comment{1D temporal U-Net, FiLM on $\mathrm{MLP}([\mathrm{emb}(\tau_i) \| \vz_i^t])$}
            \State Predict gripper logits: $\vg_i \leftarrow \phi_{\text{grip}}^i(\vz_i^t) \in \R^K$.
        \EndFor
        \State Compute joint loss: $\gL(\boldsymbol{\phi}) \leftarrow \sum_{i=1}^N \Big( \frac{1}{6K}\,\|\hat\vepsilon_i - \vepsilon_i\|_2^2 + \lambda\,\frac{1}{K}\sum_{k=1}^{K} \mathrm{BCE}(g_{i,k}, y_{i,k}) \Big)$, averaged over $\gB$.
        \State Update $\boldsymbol{\phi} \leftarrow \text{Adam}(\boldsymbol{\phi}, \nabla_{\boldsymbol{\phi}}\gL)$ with gradient clipping $\|g\|_2 \le 1.0$ and cosine learning-rate decay $10^{-4} \to 10^{-5}$.
    \EndFor
    \State Evaluate $\gL_{\text{val}}$ on $\gV_{\text{val}}$ (no proprio dropout; fixed $\tau, \vepsilon$ draws).
    \If{$\gL_{\text{val}} < \text{best\_val\_loss}$}
        \State Save optimal parameters $\boldsymbol{\phi}^* \leftarrow \boldsymbol{\phi}$; $\text{best\_val\_loss} \leftarrow \gL_{\text{val}}$.
    \ElsIf{epochs without validation improvement $> 25$}
        \State \textbf{break} (Early stopping triggered).
    \EndIf
\EndFor
\State \Return $\boldsymbol{\phi}^*$
\end{algorithmic}
\end{algorithm}

\newpage

\section{Detailed Architectural Parameters \& Training Budget}
\label{app:budget}

\subsection{CoLA Specifications and Hyperparameters}
\label{app:budget_hyperparams}

\begin{table}[ht]
\caption{\textbf{CoLA specifications and hyperparameters.} Parameter counts are exact, read from the trained checkpoints; times are on one NVIDIA A100-SXM4-40GB. Channel $= \phi_{\text{enc}} + \phi_{\text{dec}}$; the proprioceptive encoder ($4{,}672$ parameters per arm) is counted with the action head. Percentages are relative to the frozen backbone ($93$M Octo-Base, ${\sim}3.3$B $\pi_{0.5}$).}
\label{tab:system_params}
\begin{center}
\small
\begin{tabular}{lp{0.64\linewidth}}
\toprule
\textbf{Component} & \textbf{Value} \\
\midrule
Backbone feature ($\vf_i$) & Octo-Base \texttt{readout\_action} token, $d = 768$, frozen \\
Proprio encoder ($\phi_{\text{prop}}$) & Dense$(7 \to 64)$--ReLU--Dense$(64 \to 64)$; dropout $p = 0.1$ in training \\
Self representation ($\vh_i$) & $\R^{832}$ (wrist-only); $\R^{1600}$ with the overhead feature \\
$\pi_{0.5}$ state input & Own $7$-d state as prompt tokens (CoLA and LoRA alike); CoLA also via $\phi_{\text{prop}}$ \\
\midrule
Message encoder ($\phi_{\text{enc}}$) & Dense$(832 \to 96)$--ReLU--Dense$(96 \to 64)$, $d_m = 64$ (input $1600$ with overhead) \\
Message decoder ($\phi_{\text{dec}}$) & Dense$(64 \to 96)$--ReLU--Dense$(96 \to 768)$, one per ordered pair \\
\midrule
Joint denoiser ($\phi_{\text{unet}}$) & 1D U-Net, channels $(128, 256)$, kernel $5$, GroupNorm$(8)$, Mish; FiLM from Dense$(256)$--Mish--Dense$(256)$ on $[\mathrm{emb}_{64}(\tau) \,\|\, \vz_i]$ \\
Gripper head ($\phi_{\text{grip}}$) & Dense$(\vz_i \to 128)$--Mish--Dense$(128 \to K)$, binary logits \\
Diffusion and loss & $S = 100$, cosine schedule, $\epsilon$-prediction, $10$-step DDIM; $\epsilon$-MSE $+\ 0.25\,$BCE \\
\midrule
Trainable: 2-arm and tray, wrist & $14.47$M ($15.6\%$), channel $0.33$M ($0.36\%$) \\
Trainable: 2-arm, overhead & $15.20$M ($16.3\%$), channel $0.48$M ($0.52\%$) \\
Trainable: 3-arm, overhead & $23.93$M ($25.7\%$), channel $0.96$M ($1.04\%$) \\
Trainable: tray, $\pi_{0.5}$ & $16.93$M (${\sim}0.5\%$), channel $0.83$M (${\sim}0.03\%$) \\
\midrule
Optimisation & Adam ($\beta = (0.9, 0.999)$, $\epsilon = 10^{-8}$, no weight decay), clip $1.0$; cosine LR $10^{-4} \to 10^{-5}$ over $150$ epochs; batch $128$ \\
Model selection & Best validation loss; early stopping patience $25$ \\
Episodes (train/val/test) & 2-arm $120/15/15$; tray $120/15/15$; 3-arm $104/13/13$ \\
Compute & Feature caching ${\sim}5$--$12$ min per dataset; training ${\sim}35$ min (2-arm), ${\sim}45$ min (3-arm), ${\sim}11$ min ($\pi_{0.5}$) \\
\bottomrule
\end{tabular}
\end{center}
\end{table}

\subsection{Language Instructions}
\label{app:budget_instructions}

\begin{table}[htb!]
\centering
\small
\caption{\textbf{Language instructions.} The box is red in every task, and in the tray task only its marker changes colour, so no instruction reveals the hidden information. The role-specific instructions give the baselines more task information than CoLA; on the tray task, the $\pi_{0.5}$ baseline's Arm~B is also told that a tray exists.}
\label{tab:instructions}
\begin{tabular}{llp{0.5\linewidth}}
\toprule
\textbf{Model} & \textbf{Arm} & \textbf{Instruction} \\
\midrule
CoLA (Octo-Base and $\pi_{0.5}$), all tasks & all & \texttt{coordinate with partner} \\
Octo baselines, decentralised & A & \texttt{pick up the red box and hand it to the other arm} \\
 & B, C & \texttt{take the red box from the other arm} \\
Octo baselines, centralised & -- & \texttt{pick up the red box with one arm and hand it to the other} \\
$\pi_{0.5}$ LoRA baseline (tray) & A & \texttt{hand the box to the other arm} \\
 & B & \texttt{take the box from the other arm and put it in a tray} \\
\bottomrule
\end{tabular}
\end{table}

\clearpage

\subsection{Trainable Capacity and Training Cost}
\label{app:budget_cost}

\begin{table}[htb!]
\centering
\small
\renewcommand{\arraystretch}{1.3} 
\caption{\textbf{Trainable capacity and training cost by task and method.} Every method within a task uses the same optimiser steps at batch size $128$, which matches the \emph{optimisation} budget but not compute or capacity: CoLA trains on cached backbone features, so it does a fraction of the work per step. Trainable parameters are the modules updated by gradient descent, summed over the policies a method deploys (two for the 2-arm tasks, three for the 3-arm task); the Octo baselines' action head is U-Net at horizon $12$ ($68.5$M), and the centralised variants differ only in output width ($7 \to 14$), adding ${<}0.1$M. GPU-hours are job wall-clock on one NVIDIA A100-SXM4-40GB, summed over the policies a method deploys; CoLA trains all arms in a single model. They exclude CoLA's one-time feature extraction ($5$--$12$\,min per Octo-Base dataset, amortised over every CoLA run and ablation on that dataset). The centralised policy encodes both arms' cameras in every forward pass, which is why on the tray task it costs more than the two decentralised policies combined.}
\label{tab:training_cost}
\begin{tabular*}{\textwidth}{@{\extracolsep{\fill}}llrrr}
\toprule
\textbf{Task} & \textbf{Method} & \textbf{Steps} & \textbf{Trainable} & \textbf{GPU-h} \\
\midrule
\multicolumn{5}{l}{\emph{Occluded tray handover (wrist only), 2 arms}} \\
& CoLA (with message)            & $20{,}250$ & $14.5$M & $0.6$ \\
& CoLA (message zeroed)          & $20{,}250$ & $14.5$M & $0.6$ \\
& Decentralised, head-only       & $20{,}250$ & $137.0$M & $6.0$ \\
& Centralised, head-only         & $20{,}250$ & $68.5$M  & $6.4$ \\
& Decentralised, full fine-tune  & $20{,}250$ & $323.0$M & $6.1$ \\
& Centralised, full fine-tune    & $20{,}250$ & $161.5$M & $6.7$ \\
\midrule
\multicolumn{5}{l}{\emph{2-arm handover (overhead)}} \\
& CoLA (with message)            & $12{,}600$ & $15.2$M  & $0.6$ \\
& CoLA (message zeroed)          & $12{,}600$ & $15.2$M  & $0.6$ \\
& Decentralised, head-only       & $12{,}600$ & $137.0$M & $3.7$ \\
& Centralised, head-only         & $12{,}600$ & $68.5$M  & $2.1$ \\
& Decentralised, full fine-tune  & $12{,}600$ & $323.0$M & $3.8$ \\
& Centralised, full fine-tune    & $12{,}600$ & $161.5$M & $2.1$ \\
\midrule
\multicolumn{5}{l}{\emph{3-arm handover chain (overhead)}} \\
& CoLA (with message)            & $26{,}850$ & $23.9$M  & $0.8$ \\
& CoLA (message zeroed)          & $26{,}850$ & $23.9$M  & $0.8$ \\
& Decentralised, head-only       & $26{,}850$ & $205.5$M & $11.8^{\dagger}$ \\
& Decentralised, full fine-tune  & $26{,}850$ & $484.5$M & $12.0^{\dagger}$ \\
\midrule
\multicolumn{5}{l}{\emph{Occluded tray handover, $\pi_{0.5}$ backbone}} \\
& CoLA (with message)            & $20{,}250$ & $16.9$M  & $0.2$ \\
& Decentralised, LoRA fine-tune  & $20{,}250$ & $934.0$M & $48.8$ \\
\bottomrule
\end{tabular*}

\vspace{1mm}
\small
$^{\dagger}$ Per-arm job wall-clock: head-only $3.95$, $3.94$ and $3.94$\,h; full fine-tune $4.02$, $4.01$ and $4.01$\,h.
\end{table}

\clearpage
\section{Task Suite}
\label{app:figures}

\subsection{Qualitative Trajectory Comparison}
\label{app:figures_qual}

\begin{figure}[h]
    \centering
    \includegraphics[width=1\linewidth]{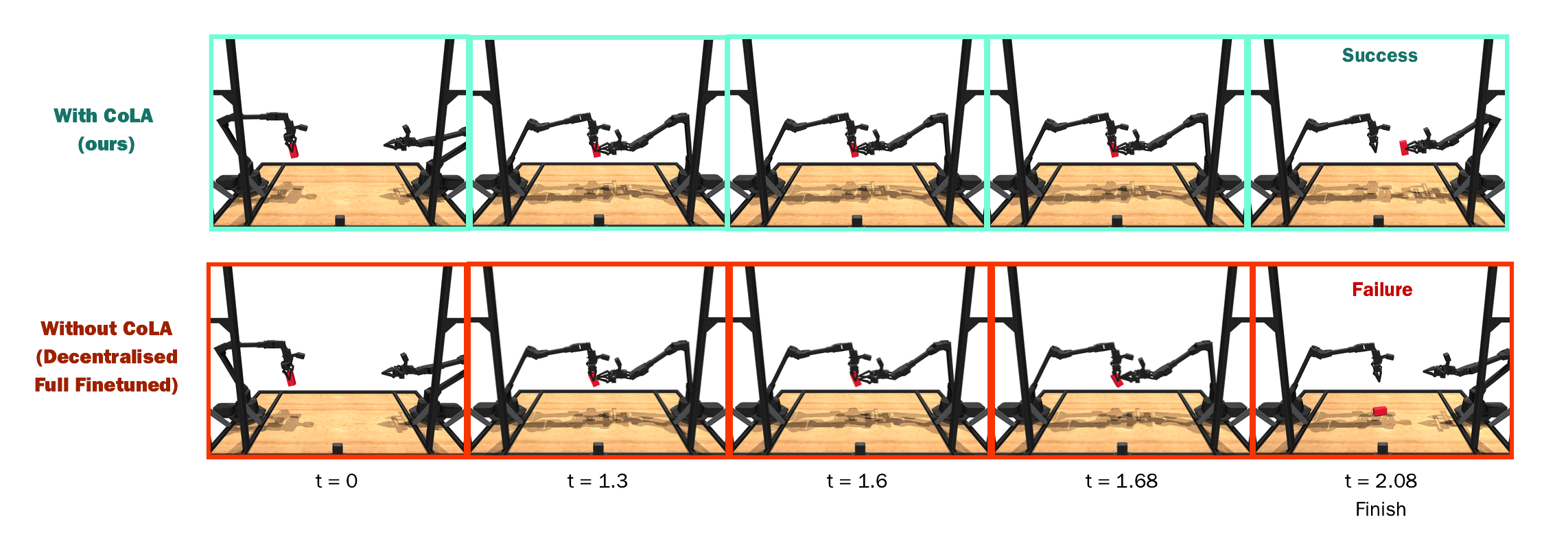}
    \caption{\textbf{Qualitative Trajectory Comparison on Handover.} Top: CoLA enables Arm A to maintain grasp stability until Arm B firmly closes its gripper. Bottom: Uncoordinated baseline desynchronises, causing premature release and dropped objects.}
    \label{fig:qual_handover}
\end{figure}

\section{Full Experiment Results}
\label{app:results}

\subsection{Two-Arm Handover Results}
\label{app:results_2arm}

\begin{table}[htb!]\centering\small\setlength{\tabcolsep}{4pt}
\caption{\textbf{2-Arm handover (overhead camera), per evaluation seed.} Success rate (\%) per 200-episode block; the seed sets the randomised start state. Octo-Base baselines use their wrist camera and the overhead camera.}\label{tab:ho_overhead_seeds}
\begin{tabular}{lcccccc}\toprule
\textbf{Method} & \textbf{0} & \textbf{1000} & \textbf{2000} & \textbf{3000} & \textbf{4000} & \textbf{Mean $\pm$ std} \\\midrule
CoLA (with message) & 88.0 & 91.0 & 91.5 & 92.5 & 90.0 & 90.6 $\pm$ 1.7 \\
CoLA (message zeroed) & 90.5 & 85.0 & 89.5 & -- & -- & 88.3 $\pm$ 2.9 \\
\midrule
Decentralised, head-only & 66.5 & 66.0 & 69.0 & 67.5 & 72.5 & 68.3 $\pm$ 2.6 \\
Centralised, head-only & 62.5 & 62.0 & 68.5 & 63.5 & 73.0 & 65.9 $\pm$ 4.7 \\
\midrule
Centralised, full fine-tune & 91.0 & 91.0 & 94.5 & 94.0 & 90.0 & 92.1 $\pm$ 2.0 \\
Decentralised, full fine-tune & 78.0 & 85.5 & 81.0 & 82.5 & 85.0 & 82.4 $\pm$ 3.1 \\
\bottomrule\end{tabular}\end{table}

\begin{table}[htb!]\centering\small\setlength{\tabcolsep}{4pt}
\caption{\textbf{2-Arm handover (wrist cameras only), per evaluation seed.} Success rate (\%) per 200-episode block at evaluation-seed offsets 0, 1000 and 2000.}\label{tab:ho_wrist_seeds}
\begin{tabular}{lcccc}\toprule
\textbf{Method} & \textbf{0} & \textbf{1000} & \textbf{2000} & \textbf{Mean $\pm$ std} \\\midrule
CoLA (with message) & 75.5 & 77.0 & 72.0 & 74.8 $\pm$ 2.6 \\
CoLA (message zeroed) & 61.5 & 59.5 & 72.0 & 64.3 $\pm$ 6.7 \\
\midrule
Decentralised, head-only & 42.0 & 43.5 & 46.0 & 43.8 $\pm$ 2.0 \\
Centralised, head-only & 19.5 & 23.5 & 20.0 & 21.0 $\pm$ 2.2 \\
\midrule
Centralised, full fine-tune & 81.5 & 81.5 & 77.0 & 80.0 $\pm$ 2.6 \\
Decentralised, full fine-tune & 71.0 & 69.5 & 67.5 & 69.3 $\pm$ 1.8 \\
\bottomrule\end{tabular}\end{table}

\subsection{Three-Arm Handover Chain Results}
\label{app:results_3arm}

\begin{table}[H]\centering\small\setlength{\tabcolsep}{4pt}
\caption{\textbf{3-Arm handover chain (overhead camera), per evaluation seed.} Success = both transfers completed and arm C holds the box for 20 steps; 200-episode blocks. CoLA rows have one block each; paired on the same 200 start states, message on vs.\ zeroed is 127 vs.\ 120 successes (exact McNemar $p = 0.50$).}\label{tab:3arm_seeds}
\begin{tabular}{lcccccc}\toprule
\textbf{Method} & \textbf{0} & \textbf{1000} & \textbf{2000} & \textbf{3000} & \textbf{4000} & \textbf{Mean $\pm$ std} \\\midrule
CoLA (with message) & 63.5 & -- & -- & -- & -- & 63.5$^\dagger$ \\
CoLA (message zeroed) & 60.0 & -- & -- & -- & -- & 60.0$^\dagger$ \\
\midrule
Decentralised, head-only & 20.0 & 16.5 & 21.0 & 21.0 & 24.0 & 20.5 $\pm$ 2.7 \\
Decentralised, full fine-tune & 36.0 & 24.0 & 33.0 & 32.0 & 33.0 & 31.6 $\pm$ 4.5 \\
\bottomrule\end{tabular}\end{table}

\begin{table}[htb!]\centering\small
\caption{\textbf{3-Arm chain, per-stage rates (\%)}, pooled over all evaluated blocks: each conditional is computed as the total over the total across blocks, not as a mean of per-block rates. A$\to$B: first transfer. B$\to$C$\mid$A$\to$B: second transfer among episodes that completed the first. Held$\mid$B$\to$C: arm C holds the box for the required 20 steps, among episodes that completed the second transfer. Success is the product of the three stages, e.g.\ $42.9 \times 51.3 \times 93.2 = 20.5$ for the head-only baseline.}\label{tab:3arm_stages}
\begin{tabular}{lcccccc}\toprule
\textbf{Method} & \textbf{Blocks} & \textbf{$n$} & \textbf{Success} & \textbf{A$\to$B} & \textbf{B$\to$C$\mid$A$\to$B} & \textbf{Held$\mid$B$\to$C} \\\midrule
CoLA (with message) & 1 & 200 & 63.5 & 79.5 & 79.9 & 100.0 \\
CoLA (message zeroed) & 1 & 200 & 60.0 & 75.0 & 80.0 & 100.0 \\
Decentralised, head-only & 5 & 1000 & 20.5 & 42.9 & 51.3 & 93.2 \\
Decentralised, full fine-tune & 5 & 1000 & 31.6 & 50.1 & 64.1 & 98.4 \\
\bottomrule\end{tabular}\end{table}

\clearpage
\subsection{Occluded Tray-Handover Results}
\label{app:results_tray}

\begin{table}[htb!]\centering\small\setlength{\tabcolsep}{3pt}
\caption{\textbf{Occluded tray task, seed-0 block.} All methods on the same 150 episodes (seed block 0, 50 per marker colour). Success = correct tray over all episodes, with 95\% Wilson interval. Handover and Reached tray are rates over all episodes; Correct$\mid$handover and Correct$\mid$tray condition on those stages (chance on Correct$\mid$tray is 33.3\%).}\label{tab:tray_blocks}
\begin{tabular}{lcccccc}\toprule
\textbf{Method} & \textbf{$n$} & \textbf{Success} & \textbf{Handover} & \textbf{Reached tray} & \textbf{Correct$\mid$handover} & \textbf{Correct$\mid$tray} \\\midrule
CoLA (with message) & 150 & 76.0 [68.6, 82.1] & 92.7 & 88.7 & 82.0 & 85.7 \\
CoLA (message zeroed) & 150 & 1.3 [0.4, 4.7] & 8.0 & 6.0 & 16.7 & 22.2 \\
\midrule
Decentralised, head-only & 150 & 22.7 [16.7, 30.0] & 88.7 & 86.0 & 25.6 & 26.4 \\
Centralised, head-only & 150 & 16.0 [11.0, 22.7] & 68.0 & 43.3 & 23.5 & 36.9 \\
\midrule
Centralised, full fine-tune & 150 & 10.7 [6.7, 16.6] & 79.3 & 10.7 & 13.4 & 100.0\\
Decentralised, full fine-tune & 150 & 12.0 [7.7, 18.2] & 94.7 & 36.7 & 12.7 & 32.7\\
\midrule
CoLA on frozen $\pi_{0.5}$ & 150 & 37.3 [30.0, 45.3] & 53.3 & 52.0 & 70.0 & 71.8\\
Decentralised, LoRA $\pi_{0.5}$ & 150 & 34.7 [27.5, 42.6] & 100.0 & 100.0 & 34.7 & 34.7\\
\bottomrule\end{tabular}\end{table}

\begin{table}[htb!]\centering\small
\caption{\textbf{CoLA ablations on the occluded tray task.} Message width: the same 150 episodes (seeds 0, 50 per marker colour) for every width. Lighting: out-of-distribution scene lighting, 30 episodes each.}\label{tab:tray_ablations}
\begin{tabular}{llrcc}\toprule
\textbf{Ablation} & \textbf{Setting} & \textbf{$n$} & \textbf{Correct} & \textbf{Correct$\mid$handover} \\\midrule
Message width $d_m$ & 16 & 150 & 73.3 & 79.7 \\
Message width $d_m$ & 32 & 150 & 42.0 & 47.4 \\
Message width $d_m$ & 64 (default) & 150 & 76.0 & 82.0 \\
Message width $d_m$ & 128 & 150 & 73.3 & 76.9 \\
\midrule
Lighting (OOD) & bright & 30 & 3.3 & 50.0 \\
Lighting (OOD) & dark & 30 & 10.0 & 15.8 \\
Lighting (OOD) & dim & 30 & 13.3 & 19.0 \\
\bottomrule\end{tabular}\end{table}

\subsection{Linear Probes of Channel Content}
\label{app:results_probes}

\begin{table}[H]\centering\small\setlength{\tabcolsep}{4pt}
\caption{\textbf{Linear probes on CoLA messages.} Tray: multinomial logistic regression to marker colour, split by episode (chance 33.3\%); frame = per-timestep accuracy, episode = majority vote. Handover: linear regression to box position. ``Zeroed'' probes the encoder of the model trained with the channel severed.}\label{tab:probes}
\begin{tabular}{llccc}\toprule
\textbf{Task / protocol} & \textbf{Representation} & \textbf{Dim} & \textbf{Frame acc.} & \textbf{Episode acc.} \\\midrule
Tray, 5-fold CV (150 ep.) & self\_a & 832 & 100.0 & 100.0 \\
Tray, 5-fold CV (150 ep.) & msg\_a & 64 & 44.2 & 73.3 \\
Tray, 5-fold CV (150 ep.) & msg\_a (zeroed model) & 64 & 39.7 & 33.3 \\
\midrule
\textbf{Handover (wrist only)} & & & \textbf{$R^2$ ($y$ / $z$)} & \textbf{RMSE cm ($y$ / $z$)} \\\midrule
Box position & self\_a & 832 & 1.000 / 1.000 & 0.04 / 0.08 \\
Box position & msg\_a & 64 & 1.000 / 0.985 & 0.12 / 0.45 \\
Box position & msg\_a (zeroed model) & 64 & 0.999 / 0.982 & 0.13 / 0.50 \\
\bottomrule\end{tabular}\end{table}

\begin{figure}[htb!]
    \centering
    \begin{minipage}{0.48\linewidth}
        \centering
        \includegraphics[width=\linewidth]{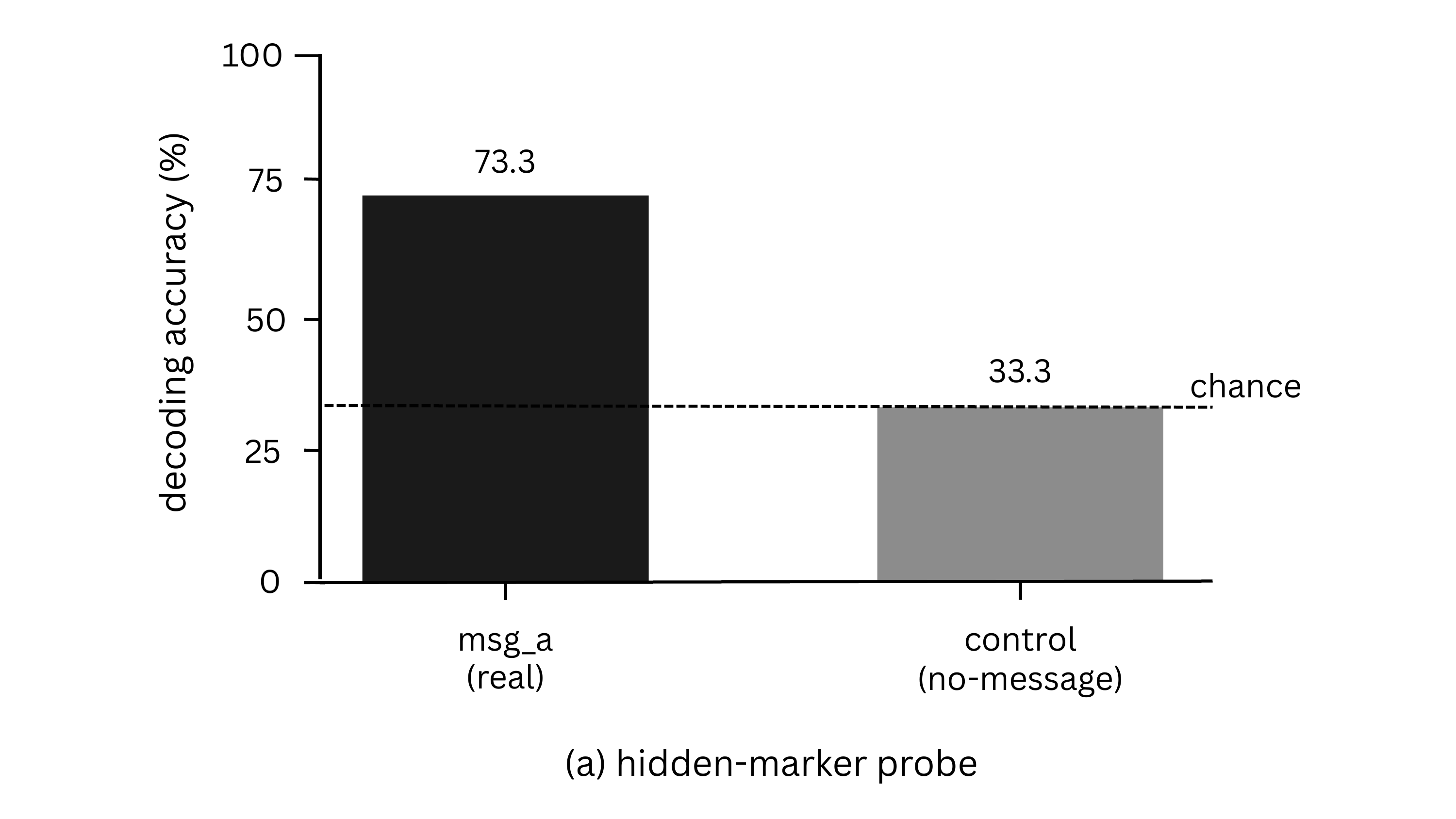}
    \end{minipage}
    \hfill
    \begin{minipage}{0.48\linewidth}
        \centering
        \includegraphics[width=\linewidth]{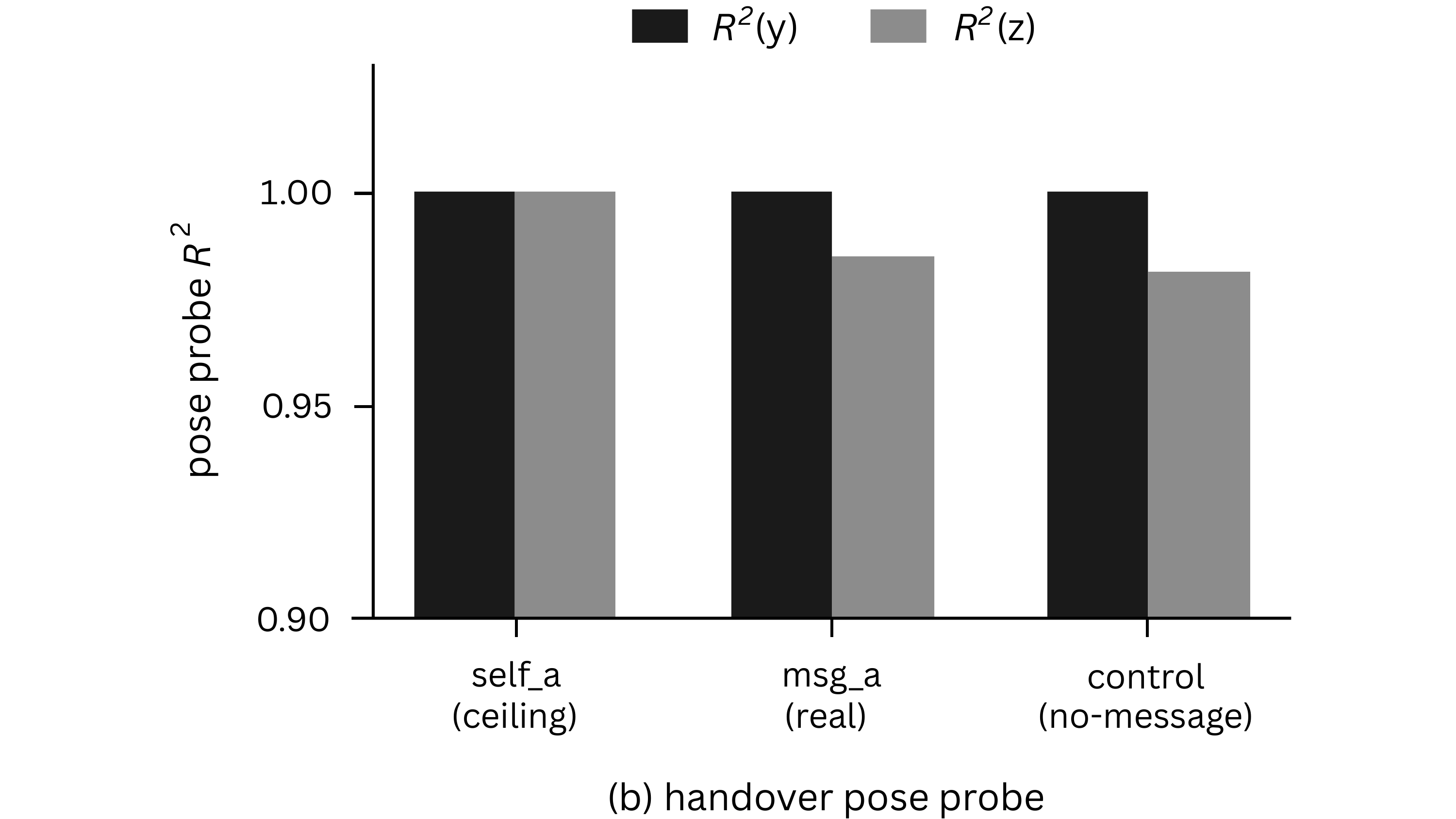}
    \end{minipage}
    \caption{\textbf{Linear probes of channel content.} \emph{Left:} on the occluded tray-handover task, the message decodes box colour well above chance while an untrained control does not, indicating channel-specific information transfer. \emph{Right:} on the wrist-only handover, the message and an untrained control both decode Arm A's presentation pose near ceiling, so pose is linearly recoverable from the raw features regardless of training, and this probe cannot isolate the channel's contribution.}
    \label{fig:probes}
\end{figure}

\clearpage
\paragraph{Pose probe.}\label{app:pose_probe}
On the wrist-only 2-arm handover, Arm A's presentation pose $(y,z)$ is decodable from self ($R^2 = 1.000\,/\,1.000$), message ($1.000\,/\,0.985$), and control ($0.999\,/\,0.982$) alike (\cref{fig:probes}, right; \cref{tab:probes}). Because Arm A rigidly grasps the object, pose is near-linear in its proprioception and survives even a random projection, so the probe cannot attribute pose to training through the channel. It does, however, show that pose is reliably present in the message, and whether it matters depends on what Arm B can already see. With the overhead view, B observes the handoff directly and zeroing the message leaves success unchanged ($90.6\%$ vs.\ $88.3\%$). With only a wrist view, the message becomes B's main source of A's pose, and zeroing it drops success from $74.8\%$ to $64.3\%$.

\subsection{Zero-Shot Pretrained Octo}
\label{app:results_zeroshot}

\begin{table}[htb!]\centering\small
\caption{\textbf{Zero-shot pretrained Octo} (no fine-tuning; Bridge action statistics; overhead camera; 200 episodes, 350 control steps). Full task: pick up and hand over. HO: episodes start with arm A holding the box.}\label{tab:zero_shot}
\begin{tabular}{llcccc}\toprule
\textbf{Backbone} & \textbf{Task} & \textbf{Success} & \textbf{Transfer} & \textbf{A touches box} & \textbf{B touches box} \\\midrule
Octo-Small & Full task & 0.0 & 0.0 & 0.0 & 0.0 \\
Octo-Small & HO & 0.0 & 0.0 & 29.0 & 0.0 \\
Octo-Base & Full task & 0.0 & 0.0 & 2.5 & 1.0 \\
Octo-Base & HO & 0.0 & 0.0 & 11.0 & 0.0 \\
\bottomrule\end{tabular}\end{table}

\subsection{Validation Gripper Accuracy}
\label{app:results_gripper}

\begin{table}[htb!]\centering\small
\caption{\textbf{Validation gripper accuracy (\%)} per arm, taken at each run's best-validation epoch, for the CoLA runs behind \cref{tab:main_results}. ---: arm not present in the task.}\label{tab:gripper_acc}
\begin{tabular}{lccc}\toprule
\textbf{Task} & \textbf{Arm A} & \textbf{Arm B} & \textbf{Arm C} \\\midrule
2-Arm handover (wrist only) & 98.8 & 98.8 & --- \\
2-Arm handover (overhead) & 98.9 & 98.7 & --- \\
Occluded tray handover & 100.0 & 100.0 & --- \\
3-Arm handover chain & 99.7 & 99.5 & 99.8 \\
\bottomrule\end{tabular}\end{table}

\clearpage
\section{Additional Training Dynamics}
\label{app:training_curves}

\begin{figure}[h]
    \centering
    \includegraphics[width=\linewidth]{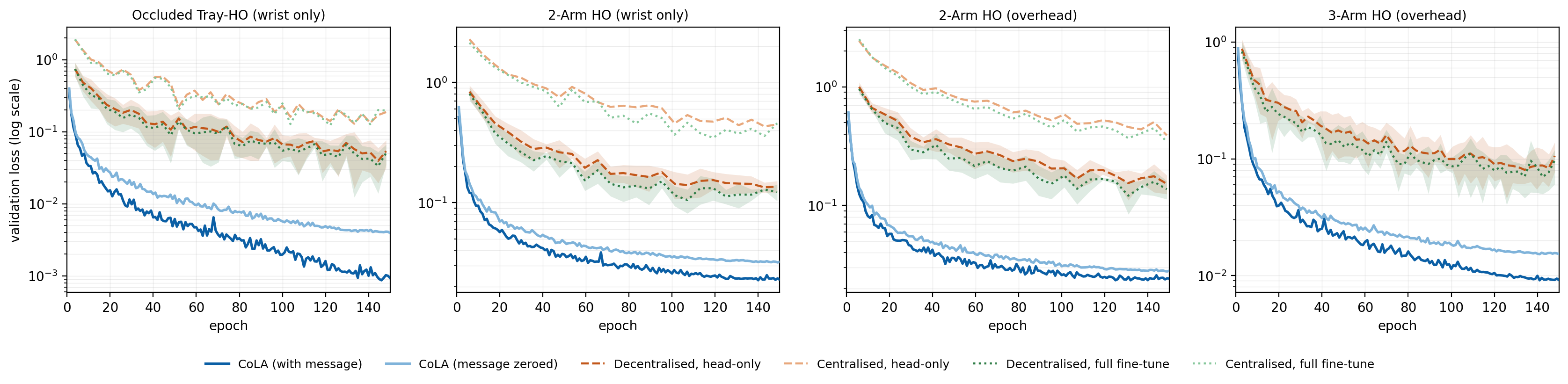}
    \caption{\textbf{Validation loss for all training runs in \cref{tab:main_results}},
    one panel per coordination regime, ordered as in the main table. Octo steps are
    converted to epochs so both families share the matched $150$-epoch budget.
    Decentralised rows train one policy per arm; the line is their mean and the shading
    their range.
    }
    \label{fig:loss_curves}
\end{figure}

\cref{fig:loss_curves} shows validation loss for every training run behind
\cref{tab:main_results}. All runs use the matched budget of $150$ epochs at
batch size $128$; the Octo baselines' optimiser steps are converted to epochs by
dividing by that task's steps per epoch ($84$, $135$ and $179$ for the two-arm,
tray and three-arm tasks). No CoLA run triggered early stopping, and the best
validation epoch ranged from $131$ to $149$ across all CoLA runs, so every run
consumed its $150$-epoch budget in full. 

Two patterns are worth noting. First, the gap between CoLA with messages and
with messages zeroed is present on every task but varies several-fold in
relative terms, and its ordering matches the behavioural results:
final validation loss is $0.0009$ vs.\ $0.0039$ on the occluded tray task
($4.3\times$), against $0.0243$ vs.\ $0.0283$ ($1.2\times$) on the two-arm
overhead task, where zeroing messages costs nothing behaviourally. Second, loss
does not rank methods the way success rate does: the centralised policies have
the \emph{highest} validation loss on every task, since fitting a $14$-D joint
action is harder than fitting a $7$-D one, yet centralised full fine-tuning
attains the best success rate on the two-arm overhead task. This is why we report
success rates rather than losses throughout.

\section{Failure Mode Taxonomy}
\label{sec:failure_modes}

Systematic inspection of failed trials identifies three orthogonal failure modes:

\subsection{Upstream Grasp Failures}
\label{sec:failure_upstream}
The initial grasping agent (Arm A) fails to secure the object due to physics edge cases during spawning. This failure is unimanual and independent of communication.

\subsection{Temporal Desynchronisation Failures}
\label{sec:failure_temporal}
Observed chiefly in the two-arm handover. Arm A ungrasps before Arm B secures contact, or Arm B closes prematurely.

\subsection{Post-Transfer Release Failures}
\label{sec:failure_posttransfer}
Specific to the three-arm chain, where success additionally requires Arm C to hold the object for $20$ steps after the second transfer. Arm C completes that transfer but loses the object during the hold in $15$ of the $220$ completed second transfers for the decentralised head-only baseline and $5$ of $321$ for decentralised full fine-tuning, and in none of CoLA's $127$. This mode is minor next to the transfers themselves: most failed three-arm episodes end at A$\to$B or B$\to$C rather than during the hold (\cref{tab:3arm_stages}).

\end{document}